\documentclass[sigconf,nonacm]{heartpreprint}

\setcopyright{none}
\AtBeginDocument{%
  }

\usepackage{times}
\usepackage{soul}
\usepackage{url}

\usepackage{caption}
\usepackage{amsmath,graphicx,hyperref}
\usepackage{amsthm}
\usepackage{booktabs}
\usepackage{algorithm}
\usepackage{algorithmic}
\usepackage[switch]{lineno}
\usepackage{lscape}
\usepackage{booktabs}
\usepackage{multirow}
\usepackage{enumitem}
\usepackage{subcaption}
\usepackage{bbm}
\usepackage[mathbold]{mathfixs}
\usepackage{cuted}

\begin{document}

\title{HEART: Hierarchical Expert Adaptive Refinement Transformer for ECG Representation Learning}

\author{Junho Song}
\affiliation{%
  \institution{MedicalAI Co., Ltd.}
  \city{Seoul}
  \country{Korea}
}
\email{jhsong@medicalai.com}
\author{Jong-Hwan Jang}
\affiliation{%
  \institution{MedicalAI Co., Ltd.}
  \city{Seoul}
  \country{Korea}
}
\email{jangood1122@medicalai.com}
\author{Joon-Myoung Kwon}
\affiliation{%
  \institution{MedicalAI Co., Ltd.}
  \city{Seoul}
  \country{Korea}
}
\email{jm.kwon@medicalai.com}
\author{Yong-Yeon Jo}
\affiliation{%
  \institution{MedicalAI Co., Ltd.}
  \city{Seoul}
  \country{Korea}
}
\email{yy.jo@medicalai.com}
\renewcommand{\shortauthors}{Song et al.}

\begin{abstract}
Learning robust representations from electrocardiograms (ECGs) is challenging because repeated cardiac cycles coexist with subtle and sparse diagnostic cues in local morphology and rhythm-level context. Existing ECG foundation models commonly rely on Transformer-based patch encoders with uniformly shared dense refinement pathways, limiting adaptive processing across heterogeneous ECG tokens. We propose \textbf{HEART (Hierarchical Expert Adaptive Refinement Transformer)}, an ECG foundation model that introduces stage-specific conditional expert refinement during self-supervised pre-training. HEART places a local-level MoE over patch-level representations before temporal contextualization and a global-level MoE over context-integrated token representations after self-attention. Token-level Top-1 routing allows different patches or contextual tokens within the same ECG to follow different expert pathways. On public ECG benchmarks, HEART achieves strong performance under both linear probing and fine-tuning, with particularly large frozen-encoder gains on PTB-XL Form and Rhythm (+6.76 AUROC points each over the strongest protocol-reproduced baseline). Routing analyses provide qualitative evidence of complementary local and global expert specialization. We release our code and pretrained models at \url{https://anonymous.4open.science/r/HEART-5E9F}.
\end{abstract}


\received{23 May 2026}

\maketitle

\section{Introduction}

Cardiovascular diseases (CVDs) remain the leading cause of mortality worldwide, highlighting the need for accurate, scalable, and robust diagnostic tools. 
The electrocardiogram (ECG), which records the electrical activity of the heart, is one of the most widely used non-invasive modalities for detecting a broad range of cardiac abnormalities~\cite{hinton2018incidence}. 
Recently, automated ECG analysis has moved beyond conventional rule-based systems and task-specific convolutional neural networks (CNNs) toward large-scale self-supervised ECG foundation models~\cite{na2024guiding,song2024foundation,mckeen2024ecg,li2024electrocardiogram}. 
These models have shown strong transferability across diverse downstream tasks by learning general-purpose ECG representations from unlabeled data~\cite{chen2020simple,kiyasseh2020clocs,wang2023contrast,na2024guiding}.

Despite these advances, ECG representation learning must account for the distinctive multi-level structure of ECG signals. 
A standard ECG recording consists of repeated cardiac cycles, yet clinically relevant abnormalities are often reflected in subtle variations of waveform morphology, temporal intervals, ST--T or QT patterns, rhythm organization, and inter-lead relationships~\cite{kligfield2007standardization,rautaharju2009standardization,wagner2009standardization}. 
These cues naturally span different representation levels: local waveform variations are expressed within short temporal segments, whereas rhythm organization and inter-lead coordination require longer-range contextual integration. 
Accordingly, effective ECG representation learning requires models that preserve fine-grained local morphology while integrating global temporal and rhythm-level context.

This local-to-global structure has motivated recent ECG foundation models to adopt Transformer-based patch encoders~\cite{na2024guiding,song2024foundation,mckeen2024ecg,li2024electrocardiogram}. 
Temporal patching provides local units that can encode short-range waveform morphology, while self-attention aggregates information across patches to form context-integrated token representations. 
However, representation refinement within Transformer blocks is still typically performed by shared dense feed-forward modules applied uniformly across tokens. 
As a result, both patch-level representations and context-integrated token representations are refined through shared transformation pathways, without explicitly adapting capacity to their different local or global characteristics. 
This shared refinement structure may limit the ability of existing models to handle heterogeneous ECG patterns across representation levels.

To address this limitation, ECG representation learning may benefit from conditional refinement mechanisms that specialize token processing according to representation-level characteristics.
Mixture-of-Experts (MoE) architectures provide a natural framework for such specialization by routing tokens to different expert modules.
Prior MoE studies have primarily focused on capacity scaling and sparse computation in large models~\cite{lepikhin2020gshard,zhou2022mixture,puigcerver2023sparse}.
However, the use of MoE as a representation-level refinement mechanism remains comparatively underexplored in physiological time-series foundation models.
We therefore focus on MoE as a stage-specific representation-refinement mechanism aligned with the hierarchy induced by patch tokenization and self-attention.

In this work, we propose \textbf{HEART (Hierarchical Expert Adaptive Refinement Transformer)}, a hierarchical ECG foundation model that treats MoE placement as a representation-learning design axis rather than as a generic scaling component. 
HEART allocates conditional expert capacity to both local and global representation stages within a self-supervised Transformer-based encoder. 
Specifically, HEART deploys MoE modules at two complementary stages:
(i) a \emph{local-level} expert refinement stage operating on patch embeddings before temporal context is integrated, and
(ii) a \emph{global-level} expert refinement stage operating on post-attention token representations after context-aware representations are formed. 
The former targets local waveform morphology and signal quality, whereas the latter targets rhythm-level dependencies. 
Token-wise Top-1 routing encourages different patches or contextual tokens within the same ECG to follow different expert pathways. 
Following prior hybrid ECG foundation modeling~\cite{song2024foundation}, HEART is pretrained with masked reconstruction and contrastive learning, augmented with auxiliary routing regularization for the MoE modules.

To validate the proposed design, we evaluate HEART on public ECG benchmarks using both linear probing and fine-tuning protocols (Tables~\ref{tab:perform_linear}--\ref{tab:perform_fine}). 
HEART achieves strong performance across tasks, with especially large linear-probing gains on PTB-XL Form and Rhythm (+6.76 AUROC points each over the strongest protocol-reproduced baseline). 
These frozen-encoder results suggest that downstream-relevant structure is already accessible to a linear classifier, while fine-tuning further improves AUPRC in several settings, including PTB-XL Form, PTB-XL Rhythm, and CPSC2018.
We further analyze expert routing and utilization patterns to examine how stage-specific MoE refinement organizes representations across local and global stages.

Our contributions are summarized as follows:
\begin{itemize}[leftmargin=*]
    \item We introduce stage-specific conditional expert refinement into self-supervised ECG representation learning, aligning local expert refinement with patch-level representations and global expert refinement with context-integrated token representations.
    \item We conduct comprehensive evaluations on public ECG benchmarks under both linear probing and fine-tuning protocols, assessing the transferability of MoE-refined representations across downstream tasks.
    \item We provide detailed analyses of expert routing and utilization, offering qualitative evidence that local- and global-level MoE modules exhibit complementary specialization patterns at different representation levels.
\end{itemize}

\begin{figure*}
    \centering
    \includegraphics[width=.9\linewidth]{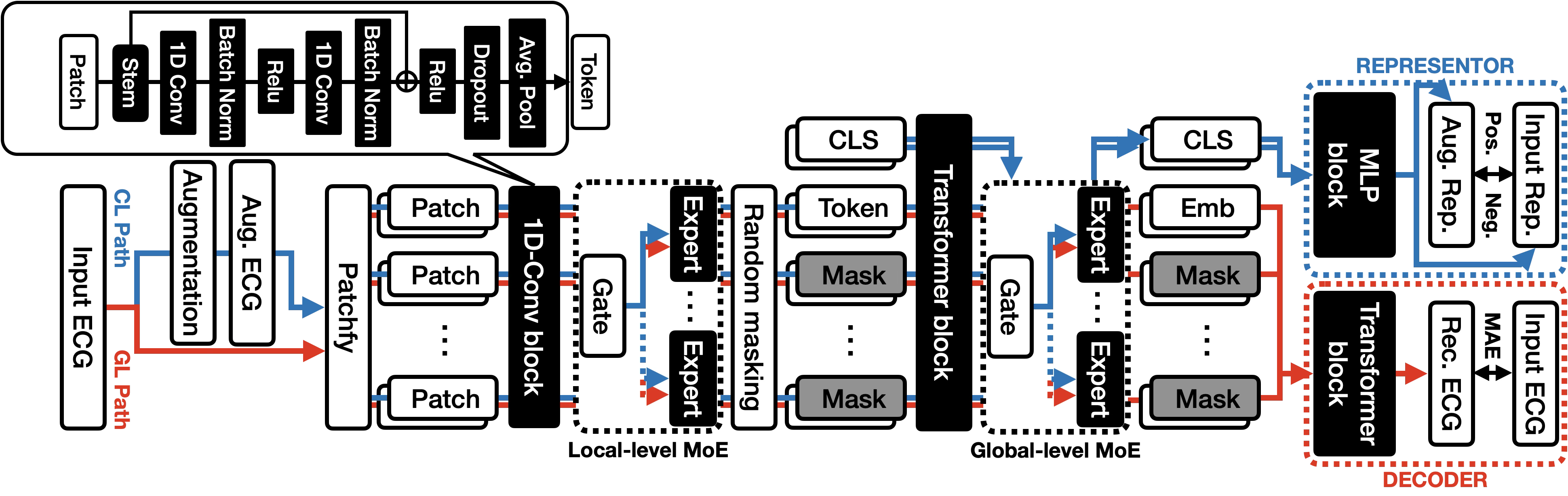}
    \vspace{-.2cm}
    \caption{
    Overview of HEART.
    Local-level MoE refines patch embeddings, and global-level MoE refines final encoder tokens.
    }
    \vspace{-.2cm}
    \Description{Overview diagram of HEART showing local and global Mixture-of-Experts modules integrated into the ECG representation learning pipeline.}
    \label{fig:overview}
\end{figure*}

\section{Related Work}

\subsection{ECG Foundation Models and Self-Supervised Representation Learning}
Self-supervised learning has become a central paradigm for ECG representation learning, supported by the availability of large-scale unlabeled ECG datasets.
Prior studies have explored contrastive learning~\cite{kiyasseh2020clocs,wang2023contrast}, masked reconstruction~\cite{na2024guiding,mckeen2024ecg,zhang2025ecgfm}, and hybrid objectives~\cite{song2024foundation} to learn transferable ECG representations.
Masked autoencoding frameworks adapted from vision~\cite{he2022masked} have been particularly effective for learning temporal continuity and waveform structure through reconstruction-based pretraining.

Recent ECG foundation models have further improved transferability by scaling pretraining corpora, strengthening encoder architectures, and incorporating richer supervision sources~\cite{mckeen2024ecg,li2024electrocardiogram,song2024foundation}.
This trend is consistent with broader time-series foundation models, such as MOMENT and MOIRAI, which demonstrate the benefit of large-scale pretraining across heterogeneous time-series datasets~\cite{goswami2024moment,woo2024moirai}.
ECG representation learning has also expanded toward multimodal ECG-text models that align ECG signals with clinical reports or language-model-derived knowledge for cross-modal reasoning and zero-shot transfer~\cite{wang2025token,liu2024zero,chen2025deri}.
While these studies have advanced ECG foundation models through improved objectives, scale, and supervision, conditional specialization within the encoder itself remains comparatively less explored.

\subsection{Multi-scale and Hierarchical ECG Modeling}
ECG analysis requires modeling structure across multiple temporal and physiological scales, including local waveform morphology, beat-level organization, rhythm-level temporal patterns, and cross-lead coordination.
Transformer architectures are well-suited for capturing long-range dependencies through self-attention~\cite{vaswani2017attention,na2024guiding,mckeen2024ecg}, and ViT-style patch tokenization~\cite{dosovitskiy2020image} has facilitated large-scale ECG modeling in recent foundation models~\cite{song2024foundation,li2024electrocardiogram}.

Several ECG-targeted approaches have incorporated multi-scale or hierarchical inductive biases, including spatio-temporal masked modeling~\cite{na2024guiding}, token-to-rhythm ECG-language supervision~\cite{wang2025token}, heartbeat-based ECG language modeling~\cite{jin2025reading}, and architectures based on convolutional branches, multi-resolution tokenization, or lead-aware attention~\cite{kan2023dg,hammer2025fusion}.
These studies show that ECG representation learning benefits from explicitly modeling both local signal morphology and higher-level temporal organization.
HEART addresses a complementary design question: rather than proposing a general replacement for ECG-specific hierarchical architectures, it studies where token-level conditional expert refinement should be introduced within a self-supervised Transformer-based ECG encoder.

\subsection{Mixture-of-Experts and Conditional Computation}
Mixture-of-Experts (MoE) architectures introduce conditional computation by routing inputs to specialized expert networks.
Early sparsely gated MoE models and subsequent large-scale variants showed that model capacity can be increased without a proportional increase in computation~\cite{shazeer2017outrageously,lepikhin2020gshard,fedus2022switch}.
Recent studies have further explored alternative routing and assignment mechanisms, including expert-choice routing and soft expert assignment~\cite{zhou2022mixture,puigcerver2023sparse}.

MoE has also been applied to representation learning and time-series foundation models.
For example, Time-MoE scales time-series forecasting models through sparse experts~\cite{shi2025timemoe}, while ECG-related studies have explored expert-style mechanisms for downstream adaptation, task-specific specialization, or model fusion~\cite{mansourian2025ecgemotionnet,xu2025enecg}.
In contrast to MoE, primarily used for scaling, forecasting, adaptation, or ensemble-style modeling, HEART uses MoE as a conditional refinement mechanism during self-supervised ECG pretraining.
By placing expert modules at both local patch-level and global context-integrated stages, HEART investigates stage-specific expert specialization over heterogeneous ECG representations.

\section{Proposed Method: HEART}
\label{sec:method}

We propose \textbf{HEART (Hierarchical Expert Adaptive Refinement Transformer)}, a self-supervised ECG foundation model that introduces stage-specific conditional expert refinement into a Transformer-based encoder.
HEART applies expert refinement at two complementary representation stages: a local-level MoE operates on patch-level representations before temporal contextualization, whereas a global-level MoE operates on context-integrated token representations after self-attention.
HEART therefore studies where token-level conditional capacity should be placed within the ECG representation hierarchy, rather than proposing a new routing algorithm or replacing ECG-specific hierarchical architectures.
An overview of HEART is shown in Fig.~\ref{fig:overview}.

\subsection{Architecture Overview}
\label{sec:architecture}

Given an input ECG $\mathbf{x} \in \mathbb{R}^{C \times T}$, where $C$ denotes the number of leads and $T$ denotes the temporal length, HEART first divides the signal into $N$ temporal patches:
\begin{equation}
    \mathbf{x} \rightarrow \{\mathbf{x}_p\}_{p=1}^{N}.
\end{equation}
Each patch is encoded into a patch-level representation:
\begin{equation}
    \mathbf{e}_p = f_{\mathrm{patch}}(\mathbf{x}_p),
    \quad
    \mathbf{e}_p \in \mathbb{R}^{D}.
\end{equation}
These patch embeddings define the local representation space before temporal context is aggregated by the Transformer encoder.

\paragraph{\textbf{Local-level expert refinement.}}
The local-level MoE operates on patch embeddings before self-attention.
For each patch representation $\mathbf{e}_p$, the router selects one local expert and produces a refined patch representation:
\begin{equation}
    \tilde{\mathbf{e}}_p
    =
    \Phi^{\mathrm{local}}_{k_p}(\mathbf{e}_p),
    \quad
    k_p = \arg\max_i s_{p,i},
\end{equation}
where $\Phi^{\mathrm{local}}_{k_p}$ denotes the selected local expert and $s_{p,i}$ is the routing probability for expert $i$.
This stage targets local waveform characteristics, such as morphology and noise patterns, before temporal context is aggregated.
Since routing is performed at the patch level, different segments within the same ECG can activate different local experts.

\paragraph{\textbf{Masking and Transformer encoding.}}
After local expert refinement, positional embeddings are added, and random masking is applied for masked autoencoding.
Only visible tokens are passed to the Transformer encoder together with a prepended [CLS] token:
\begin{equation}
    \{\mathbf{z}_{L-1}^{(t)}\}_{t=0}^{N'}
    =
    f_{\mathrm{enc}}
    \left(
    [\mathbf{e}_{\mathrm{CLS}}; \{\hat{\mathbf{e}}_p\}_{p=1}^{N'}]
    \right),
\end{equation}
where $t=0$ denotes the [CLS] token, $\{\hat{\mathbf{e}}_p\}_{p=1}^{N'}$ denotes the visible locally processed patch tokens, and $N'$ is the number of visible patches.
The masked positions are not provided to the encoder and are reconstructed by the decoder during pre-training.

\paragraph{\textbf{Global-level expert refinement.}}
The global-level MoE operates on post-attention token representations.
At this stage, each token has incorporated temporal and cross-lead context through the Transformer encoder.
In our implementation, global expert refinement is applied as a residual update:
\begin{equation}
    \mathbf{z}_{L}^{(t)}
    =
    \mathbf{z}_{L-1}^{(t)}
    +
    \Phi^{\mathrm{global}}_{k_t}
    \left(
    \mathbf{z}_{L-1}^{(t)}
    \right),
    \quad
    k_t = \arg\max_i s_{t,i}.
\end{equation}
This stage targets context-integrated representations associated with rhythm-level dependencies, distributed temporal patterns, and cross-lead coordination.
The [CLS] token participates in global-level routing and serves as the sequence-level representation for downstream classification, but it is not used as a reconstruction target.

\paragraph{\textbf{MAE-style decoder.}}
For masked reconstruction, after removing the [CLS] token from the encoder output, the visible token representations are projected to the decoder dimension, combined with learnable mask tokens, and reordered to match the original patch sequence.
The reordered tokens are processed by Transformer decoder blocks, followed by a linear prediction head that reconstructs each ECG patch.
The reconstruction loss is computed only over masked patches, and the decoder is discarded during downstream evaluation.

\subsection{Unified Token Routing}
\label{sec:moe_mechanism}

Both local- and global-level MoE modules use the same Top-1 token routing mechanism.
Given a token representation, the router computes expert probabilities and dispatches the token to the expert with the highest probability:
\begin{equation}
    k = \arg\max_i s_i.
\end{equation}
A capacity constraint is applied to prevent expert overloading.
For a batch of $N$ routed tokens and $E$ experts, each expert can process at most
\begin{equation}
    C
    =
    \left\lceil
    \frac{N}{E}
    \cdot
    \mathrm{capacity\_factor}
    \right\rceil
\end{equation}
tokens.
Tokens exceeding the capacity of their selected expert are not dispatched to that expert.
This constraint encourages balanced use of experts during pre-training.

\paragraph{\textbf{Rationale for Top-1 token routing.}}
We use Top-1 routing as a scoped design choice for clear token-level specialization rather than as a claim of superiority over alternative routing strategies.
For ECG representation learning, this assigns each local or contextual token to a single expert pathway while still allowing a single ECG recording to activate multiple experts across its sequence.
This is important because many ECG segments may contain recurring beat patterns, whereas diagnostically relevant variations can appear sparsely in specific patches or context-dependent tokens.
Thus, Top-1 routing encourages sharper expert specialization at the token level without imposing a one-expert-per-sample partition.

Detailed routing equations, including noisy routing, load balancing, entropy regularization, and router logit regularization, are provided in Appendix~\ref{app:moe_routing}.

\subsection{Training Objective}
\label{sec:objectives}

HEART is trained with generative, contrastive, and auxiliary routing objectives during self-supervised pre-training.
The generative and contrastive objectives follow standard formulations from prior hybrid ECG foundation modeling~\cite{song2024foundation}; therefore, we briefly summarize their roles and focus on the routing-specific objective introduced by the MoE modules.

\paragraph{\textbf{Generative and contrastive objectives.}}
The generative objective follows masked autoencoding~\cite{he2022masked}: the decoder reconstructs masked ECG patches from visible tokens and learned mask tokens, and the reconstruction loss is computed over masked positions.
This objective encourages the encoder to preserve local waveform structure and temporal continuity.

The contrastive objective follows an InfoNCE-style formulation~\cite{chen2020simple,song2024foundation}.
It is applied to sequence-level [CLS] representations obtained from different ECG views, encouraging globally discriminative ECG embeddings.
Together, the generative and contrastive objectives provide complementary supervision for local waveform reconstruction and global representation separability.

\paragraph{\textbf{Auxiliary routing objective.}}
The MoE modules introduce Top-1 routing, where each token is assigned to a single expert under a capacity constraint.
To stabilize routing and reduce expert collapse, HEART uses an auxiliary routing objective consisting of load balancing, routing entropy, and router logit regularization:
\begin{equation}
    \mathcal{L}_{\mathrm{aux}}
    =
    \mathcal{L}_{\mathrm{bal}}
    -
    \alpha
    \mathcal{L}_{\mathrm{ent}}
    +
    \lambda_z
    \mathcal{L}_{z},
\end{equation}
where $\mathcal{L}_{\mathrm{bal}}$ encourages distributed expert utilization, $\mathcal{L}_{\mathrm{ent}}$ promotes routing exploration, and $\mathcal{L}_{z}$ regularizes the magnitude of router logits.
The same auxiliary loss form is applied to both local- and global-level MoE modules during pre-training.

\paragraph{\textbf{Total pre-training objective.}}
The final pre-training objective is
\begin{equation}
\label{eq:total_loss}
    \mathcal{L}_{\mathrm{total}}
    =
    \mathcal{L}_{\mathrm{gen}}
    +
    \lambda_{\mathrm{cont}}
    \mathcal{L}_{\mathrm{cont}}
    +
    \lambda_{\mathrm{aux}}
    \left(
    \mathcal{L}_{\mathrm{aux}}^{\mathrm{local}}
    +
    \mathcal{L}_{\mathrm{aux}}^{\mathrm{global}}
    \right),
\end{equation}
where $\mathcal{L}_{\mathrm{gen}}$ and $\mathcal{L}_{\mathrm{cont}}$ denote the masked reconstruction and contrastive objectives, respectively.
The coefficients $\lambda_{\mathrm{cont}}$ and $\lambda_{\mathrm{aux}}$ control the contributions of contrastive learning and routing regularization.
The decoder and auxiliary routing losses are used only during self-supervised pre-training.
During downstream evaluation, HEART retains the pretrained encoder, including the local- and global-level MoE modules, and uses a conventional dense classifier head on the final [CLS] representation.

\begin{table*}[t]
\footnotesize
\centering
\caption{Linear-probing performance on PTB-XL and CPSC2018 (AUROC; AUPRC in parentheses where available).
Linear classifiers are trained on frozen encoders; PTB-XL Average is the mean over four diagnostic subsets.
Non-daggered baselines are protocol-reproduced under our setup.
$^\dagger$ marks external same-dataset references included only for context and not treated as protocol-matched comparisons.
Bold and underline indicate the best and second-best results among protocol-reproduced methods only.
CREMA-Max is a parameter-comparable dense CREMA variant.}
\label{tab:perform_linear}
\begin{tabular}{@{}c|ccccccccc|c@{}}
\toprule
{\textbf{Dataset}} & 
\textbf{\begin{tabular}[c]{@{}c@{}} Random Init. \end{tabular}} & 
\textbf{\begin{tabular}[c]{@{}c@{}} ECGFounder$^\dagger$ \end{tabular}} & 
\textbf{\begin{tabular}[c]{@{}c@{}}SimCLR \end{tabular}} & 
\textbf{\begin{tabular}[c]{@{}c@{}}CLoCS \end{tabular}} & 
\textbf{\begin{tabular}[c]{@{}c@{}}COMET \end{tabular}} & 
\textbf{\begin{tabular}[c]{@{}c@{}}ST-MEM \end{tabular}} & 
\textbf{\begin{tabular}[c]{@{}c@{}}MAE \end{tabular}} & 
\textbf{\begin{tabular}[c]{@{}c@{}}CREMA \end{tabular}} & 
\textbf{\begin{tabular}[c]{@{}c@{}}CREMA-Max \end{tabular}} & 
\textbf{\begin{tabular}[c]{@{}c@{}}HEART\end{tabular}} \\ 
\midrule
\textbf{PTB-XL Super} & 86.76 & 91.20 & 89.24 & 89.72 & 90.44 & 81.49 & 87.55 & \underline{91.25} & 91.05 & \textbf{92.48 (81.73)} \\ 
\textbf{PTB-XL Sub} & 85.40 & 88.03 & 87.13 & 90.43 & 90.57 & 78.52 & 88.10 & 91.37 & \underline{92.06} & \textbf{93.96 (54.04)} \\ 
\textbf{PTB-XL Form} & 63.05 & 82.52 & 74.75 & 77.49 & 75.01 & 66.25 & 77.39 & 80.14 & \underline{81.58} & \textbf{88.34 (32.68)} \\ 
\textbf{PTB-XL Rhythm} & 86.08 & 93.61 & 83.07 & 86.34 & 86.00 & 82.53 & 87.07 & \underline{88.92} & 87.18 & \textbf{95.68 (55.40)} \\ 
\midrule
\textbf{PTB-XL Average} & 80.32 & 88.84 & 83.55 & 86.00 & 85.51 & 77.20 & 85.03 & 87.92 & \underline{87.96} & \textbf{92.62 (55.96)} \\
\textbf{CPSC2018} & 74.30 & 83.18 & 87.03 & 90.24 & 90.28 & 79.05 & 87.72 & \underline{92.78} & 92.16 & \textbf{96.50 (83.03)} \\ 
\bottomrule
\end{tabular}
\end{table*}

\begin{table*}[t]
\footnotesize
\centering
\caption{Fine-tuning performance on PTB-XL and CPSC2018 (AUROC; AUPRC in parentheses where available).
For protocol-reproduced models, all parameters are optimized end-to-end.
$^\dagger$ marks external same-dataset references included only for context and not treated as protocol-matched comparisons.
Bold and underline indicate the best and second-best results among protocol-reproduced methods only.}
\label{tab:perform_fine}
\begin{tabular}{@{}c|cccccccccc|c@{}}
\toprule
\textbf{Dataset} & 
\textbf{\begin{tabular}[c]{@{}c@{}}ECGFounder$^\dagger$ \end{tabular}} &  
\textbf{\begin{tabular}[c]{@{}c@{}}SimCLR \end{tabular}} & 
\textbf{\begin{tabular}[c]{@{}c@{}}CLoCS \end{tabular}} & 
\textbf{\begin{tabular}[c]{@{}c@{}}COMET \end{tabular}} & 
\textbf{\begin{tabular}[c]{@{}c@{}}ECGFM$^\dagger$ \end{tabular}} & 
\textbf{\begin{tabular}[c]{@{}c@{}}ECG-FM$^\dagger$ \end{tabular}} & 
\textbf{\begin{tabular}[c]{@{}c@{}}ST-MEM \end{tabular}} & 
\textbf{\begin{tabular}[c]{@{}c@{}}MAE \end{tabular}} & 
\textbf{\begin{tabular}[c]{@{}c@{}}CREMA \end{tabular}} & 
\textbf{\begin{tabular}[c]{@{}c@{}}CREMA-Max \end{tabular}} & 
\textbf{\begin{tabular}[c]{@{}c@{}}HEART \end{tabular}} \\ 
\midrule
\textbf{PTB-XL Super} & - & 92.28 & 92.09 & 92.30 & - & - & 87.95 & 92.23 & \textbf{92.86} & 91.91 & \underline{92.53 (82.77)} \\ 
\textbf{PTB-XL Sub} & - & 92.04 & 90.78 & 92.20 & - & - & 87.84 & 91.12 & \underline{93.35} & 92.43 & \textbf{93.42 (51.90)} \\ 
\textbf{PTB-XL Form} & - & 84.66 & 80.43 & 78.98 & - & - & 72.98 & 85.81 & \underline{87.07} & 83.70 & \textbf{88.40 (37.32)} \\ 
\textbf{PTB-XL Rhythm} & - & 91.04 & 91.14 & 91.88 & - & - & 90.68 & 91.29 & \underline{91.94} & 90.48 & \textbf{95.09 (60.20)} \\ 
\cmidrule{1-12} 
\textbf{PTB-XL Average} & 89.77 & 90.01 & 88.84 & 88.61 & 88.85 & 90.29 & 84.86 & 90.11 & \underline{91.31} & 89.63 & \textbf{92.36 (58.05)} \\ 
\textbf{CPSC2018} & - & 94.33 & 93.87 & 94.21 & 91.91 & - & 93.20 & 95.04 & \underline{95.77} & 95.05 & \textbf{97.34 (86.17)} \\ 
\bottomrule
\end{tabular}
\end{table*}

\section{Evaluation}

\subsection{Datasets}

\paragraph{\textbf{Pre-training.}}
We pre-trained HEART using five large-scale ECG datasets available for research use: MIMIC~\cite{johnson2016mimic}, CODE15~\cite{ribeiro_2021_4916206}, UK Biobank~\cite{sudlow2015uk}, SAMI~\cite{ribeiro_2021_4905618}, and IKEM~\cite{michal_sejak_2023_8393007}.
In total, the pre-training corpus comprises approximately 1.29 million ECG recordings from 442,000 patients, covering heterogeneous ECG sources across multiple cohorts and acquisition settings.

\paragraph{\textbf{Downstream.}}
For downstream evaluation, we use PTB-XL~\cite{wagner2020ptb}, which contains approximately 21,000 ECGs annotated across four diagnostic subsets (Super, Sub, Form, and Rhythm), and CPSC2018~\cite{liu2018open}, which includes approximately 6,900 ECGs annotated with nine diagnostic labels.
For PTB-XL, we follow the official split protocol using folds 1--8 for training, fold 9 for validation, and fold 10 for testing.
For CPSC2018, we follow the split protocol used in prior ECG foundation model studies~\cite{song2024foundation}.
The same downstream splits are used for HEART and all protocol-reproduced baselines.
PTB-XL and CPSC2018 are not included in the pre-training corpus, and no downstream evaluation records are used during pre-training.

\paragraph{\textbf{Preprocessing.}}
All ECG signals used for both pre-training and downstream evaluation are resampled to 250\,Hz and truncated or padded to a fixed duration of 10 seconds.
The same preprocessing pipeline is applied to HEART and all protocol-reproduced baselines.

\subsection{Implementation Details}

We implement HEART in PyTorch.
Unless otherwise specified, HEART is pre-trained with a batch size of 256 using AdamW ($\beta_1=0.9$, $\beta_2=0.999$), an initial learning rate of $1\times10^{-4}$, linear warmup, and cosine decay.
The contrastive loss weight is set to $\lambda_{\mathrm{cont}}=0.1$, while module-specific auxiliary routing weights are selected by hyperparameter optimization.
During pre-training, MoE optimization is stabilized using capacity constraints, auxiliary routing regularization, input jitter, and noisy gating.
Detailed architectural and optimization hyperparameters are provided in Appendix~\ref{app:implementation}.

For downstream evaluation, the pre-training decoder is discarded, and the final [CLS] representation is passed to a conventional dense classifier.
The pretrained local- and global-level MoE modules are retained as part of the downstream encoder.
Auxiliary routing losses are used during pre-training and fine-tuning, but not for linear probing.
In linear probing, the pretrained encoder, including the MoE routers and experts, is frozen, and only the classifier is trained.
In fine-tuning, the encoder and classifier are updated end-to-end.

\subsection{Baselines and Comparison Protocol}

We separate baseline results into protocol-reproduced baselines and external reference results.
For protocol-reproduced baselines, we re-evaluate SimCLR, CLoCS, COMET, ST-MEM, MAE, and CREMA using the available code, following our preprocessing and downstream evaluation protocol when possible~\cite{na2024guiding,he2022masked,chen2020simple,kiyasseh2020clocs,wang2023contrast,song2024foundation}.
These baselines are used as the primary basis for direct comparison.

For methods whose training code or protocol is not publicly available and directly reproducible in our setting, including ECGFounder, ECGFM, and ECG-FM, we report results from the original publications or public benchmark releases, when available, on the same downstream dataset~\cite{li2024electrocardiogram,mckeen2024ecg,zhang2025ecgfm,almasud2026benchmarking}.
Because these results may differ due to label aggregation, data splits, preprocessing, signal windows, or adaptation strategies, they are treated as contextual references rather than protocol-matched baselines.
For PTB-XL, this distinction is particularly important because our tables report the four official diagnostic subsets, whereas some external baselines report a single aggregate over the full fine-grained PTB-XL label space.

We additionally construct CREMA-Max as a dense-capacity control.
CREMA-Max follows the same pre-training corpus and downstream evaluation protocol as CREMA and HEART, while increasing the encoder dimension from $1024$ to $1488$, the encoder depth from $4$ to $13$, and the pre-training decoder depth from $2$ to $4$.
This baseline assesses whether HEART's improvements are attributable to conditional expert routing rather than dense parameter scaling alone.

\subsection{Results}

\paragraph{\textbf{Linear probing.}}
Table~\ref{tab:perform_linear} reports linear-probing performance on PTB-XL and CPSC2018.
Linear probing evaluates the quality of pretrained representations by training a linear classifier on top of a frozen encoder.

Across all PTB-XL diagnostic subsets, HEART achieves the highest AUROC among protocol-reproduced methods.
Compared with contrastive baselines such as SimCLR, CLoCS, and COMET, and reconstruction-based baselines such as ST-MEM and MAE, HEART shows consistently stronger frozen-encoder performance.
The largest gains appear on PTB-XL Form and Rhythm, where HEART improves over the strongest protocol-reproduced baseline by +6.76 AUROC points in both cases, corresponding to relative improvements of +8.29\% and +7.60\%, respectively.
These two subsets are closely related to waveform-form characteristics and rhythm-level temporal organization, making the large margins consistent with HEART's stage-specific refinement of local and context-integrated representations.

On average, HEART reaches 92.62 AUROC across the four PTB-XL subsets, exceeding the strongest protocol-reproduced baseline by +4.66 AUROC points (+5.30\% relative).
On CPSC2018, HEART also achieves the highest linear-probing AUROC, reaching 96.50 and improving over the strongest protocol-reproduced baseline by +3.72 AUROC points (+4.01\% relative).

Overall, these results suggest that HEART learns representations in which downstream-relevant structure is already accessible to a linear classifier.

\paragraph{\textbf{Fine-tuning.}}
Table~\ref{tab:perform_fine} reports fine-tuning results, where all parameters are optimized end-to-end for protocol-reproduced models.
Under fine-tuning, several baselines improve substantially, and CREMA remains a strong dense baseline.
Nevertheless, HEART achieves the highest AUROC on PTB-XL Sub, Form, Rhythm, and the four-task average, while remaining competitive on Super.
On CPSC2018, HEART also achieves the highest AUROC among protocol-reproduced baselines, reaching 97.34.

Fine-tuning mainly improves HEART's precision--recall behavior in several settings.
Compared with linear probing, HEART increases AUPRC on PTB-XL Form from 32.68 to 37.32 (+4.64; +14.20\% relative), on PTB-XL Rhythm from 55.40 to 60.20 (+4.80; +8.66\% relative), and on CPSC2018 from 83.03 to 86.17 (+3.14; +3.78\% relative).
The PTB-XL four-task average AUPRC also increases from 55.96 to 58.05 (+2.09; +3.74\% relative).
These results suggest that end-to-end adaptation can further refine precision--recall trade-offs, especially for tasks where positive-label identification is more sensitive to classifier calibration and task-specific adaptation.

At the same time, HEART's fine-tuned AUROC remains close to its linear-probing AUROC.
This pattern suggests that the pretrained encoder already provides representations in which downstream-relevant structure is accessible to a simple classifier, while fine-tuning further adjusts the decision boundary and precision--recall trade-off for several tasks.

\paragraph{\textbf{Dense capacity control.}}
To test whether HEART's gains can be explained by dense parameter scaling alone, we compare against CREMA-Max, a parameter-comparable dense CREMA variant.
Across both linear probing and fine-tuning, CREMA-Max does not consistently improve over the original CREMA, whereas HEART achieves stronger average performance.
This suggests that the observed gains are more closely associated with stage-specific conditional expert refinement than with uniformly increasing dense model capacity.

\subsection{Ablation Studies}

We conduct ablation studies on CPSC2018 to examine how individual design components in HEART affect representation quality.
We use CPSC2018 as a representative controlled setting to isolate the effect of MoE placement under the same data splits and training protocols as the main experiments.

\paragraph{\textbf{Effect of MoE placement.}}

\begin{table}[!htbp]
\vspace{-.2cm}
\centering
\footnotesize
\caption{MoE placement ablation on CPSC2018 under fine-tuning.
CREMA is the dense backbone.
Local/Global variants add one MoE at the patch-embedding or context-integrated encoder stage, respectively.
HEART uses both local- and global-level MoE modules.
Sensitivity is evaluated at specificity = $0.98$.}
\label{tab:ablation}
\vspace{-.2cm}
\begin{tabular}{@{}c|cccc@{}}
\toprule
\textbf{CPSC2018} & \textbf{AUROC} & \textbf{AUPRC} & \textbf{F1-score} & \textbf{Sensitivity (SPEC=0.98)} \\ 
\midrule
\textbf{CREMA} & 95.77 & 81.15 & 76.67 & 76.44 \\
\textbf{Local-level MoE} & 96.30 & 85.34 & 79.26 & 77.45 \\
\textbf{Global-level MoE} & 96.04 & 83.59 & 77.83 & 76.78 \\
\textbf{HEART} & \textbf{97.34} & \textbf{86.17} & \textbf{80.55} & \textbf{79.67} \\
\bottomrule
\end{tabular}
\vspace{-.2cm}
\end{table}

Table~\ref{tab:ablation} shows that adding either local- or global-level MoE improves over the dense CREMA backbone.
The full HEART configuration achieves the best performance across all metrics, supporting the benefit of combining local patch-level transformation with global context-aware refinement.
Under the selected expert settings, the local-only variant shows slightly larger gains than the global-only variant, indicating that patch-level conditional transformation is an important contributor to the observed improvement.

\begin{figure}[!htbp]
    \vspace{-.2cm}
    \centering
    \includegraphics[width=.95\linewidth]{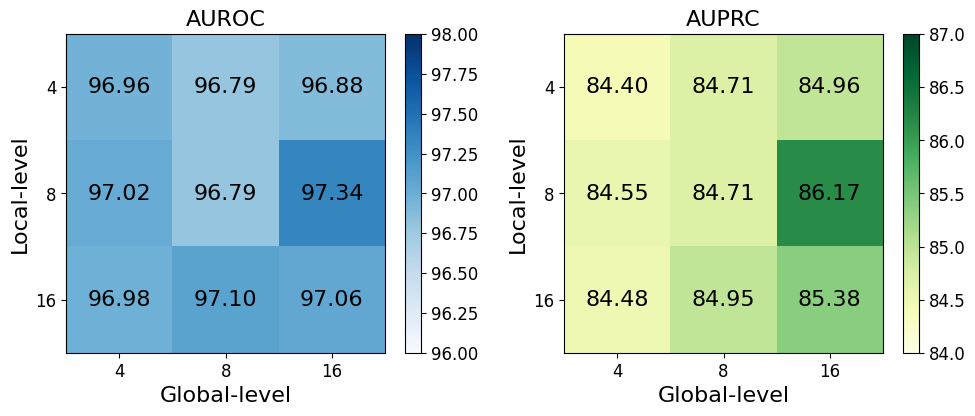}
    \vspace{-.2cm}
    \caption{
    Expert-count sensitivity on CPSC2018.
    Cells sweep local/global expert counts with all other settings fixed; AUROC (left) is stable, while AUPRC (right) is more capacity-sensitive.
    }
    \vspace{-.2cm}
    \Description{Heatmap-style performance plots showing AUROC and AUPRC for different local and global expert counts on CPSC2018.}
    \label{fig:ablation_expert_num}
\end{figure}

\paragraph{\textbf{Sensitivity to the number of experts.}}
We examine the sensitivity of HEART to the number of experts by varying local- and global-level expert counts while keeping all other settings fixed.
Figure~\ref{fig:ablation_expert_num} reports AUROC and AUPRC for combinations of $\{4,8,16\}$ experts at each level.

Across configurations, AUROC varies within a narrow range, indicating that ranking-based discrimination is relatively robust to expert count.
In contrast, AUPRC shows greater sensitivity to expert capacity.
In particular, configurations with more global-level experts tend to improve AUPRC on CPSC2018, suggesting that additional context-aware expert capacity can help improve precision--recall performance in imbalanced settings.
Based on these observations, we select the configuration that provides the best overall performance while maintaining reasonable model complexity for the main experiments.

\paragraph{\textbf{Model efficiency and performance trade-off.}}

\begin{table}[!htbp]
    \vspace{-.2cm}
    \centering
    \footnotesize
    \caption{Model size and inference cost for a 10-second 12-lead ECG.
    AUROC is PTB-XL Average under linear probing; $\Delta$ denotes the AUROC gain over CREMA.
    Parameter counts follow the pretrained model configuration; latency uses encoder-forward inference.}
    \label{tab:efficiency}
    \vspace{-.2cm}
    \begin{tabular}{@{}lccccc@{}}
    \toprule
    \textbf{Model} &
    \textbf{Params (M)} &
    \textbf{Size (MB)} &
    \textbf{Inference (ms)} &
    \textbf{AUROC} &
    \textbf{$\Delta$} \\
    \midrule
    \textbf{CREMA} & 22.09 & 96.34 & 3.26 & 87.92 & +0.00 \\
    \textbf{CREMA-Max} & 177.60 & 677.50 & 13.22 & 87.96 & +0.04 \\
    \textbf{HEART} & 161.83 & 633.52 & 9.33 & 92.62 & +4.70 \\
    \bottomrule
    \end{tabular}
    \vspace{-.2cm}
\end{table}

Table~\ref{tab:efficiency} reports model size, inference latency, and representation quality under the PTB-XL Average linear-probing setting.
Latency is measured as encoder forward time with batch size 1 on a single GPU using mixed precision, averaged over 10 runs after warm-up.
Compared with the original CREMA, HEART requires more parameters and higher latency, but provides a substantially larger representation gain.
Compared with CREMA-Max, HEART achieves substantially higher AUROC with lower latency, suggesting that conditional expert capacity is more effective than uniformly scaling the dense encoder in this setting.
Overall, this comparison supports that HEART's representation gain is unlikely to be explained by parameter count alone.

\subsection{Qualitative Analysis}
\label{sec:qualitative}

To complement the quantitative results, we analyze HEART under the PTB-XL Rhythm linear-probing setting.
The frozen-encoder nature of linear probing allows us to examine the representation structure and routing behavior learned during self-supervised pre-training.
We focus on the Rhythm subset as it contains temporally organized ECG patterns, making it suitable for analyzing context-integrated representations and expert utilization.
The following analyses provide qualitative diagnostics of the pretrained encoder and its stage-specific routing behavior.

\paragraph{\textbf{Representation space.}}
\begin{figure}[!htbp]
    \vspace{-.2cm}
    \centering
    \includegraphics[width=.99\linewidth]{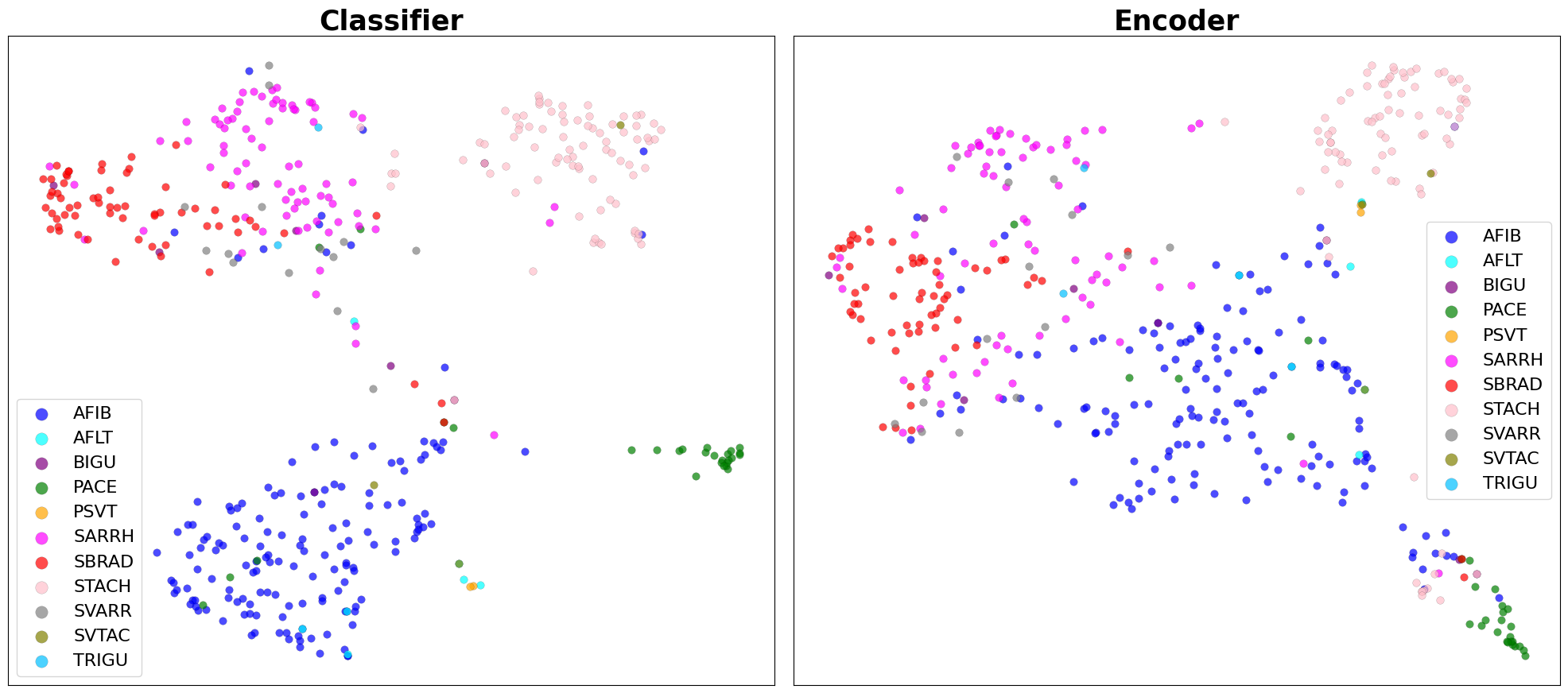}
    \vspace{-.2cm}
    \caption{
    UMAP visualization of frozen HEART [CLS] embeddings and the output space of the trained linear-probing classifier on PTB-XL Rhythm.
    }
    \vspace{-.2cm}
    \Description{UMAP scatter plots comparing frozen encoder CLS representations and the output space of the trained linear-probing classifier for PTB-XL rhythm classes.}
    \label{fig:umap_class_encode}
\end{figure}

We first examine whether rhythm-related structure is already present in the pretrained representation.
Figure~\ref{fig:umap_class_encode} compares UMAP visualizations of frozen encoder [CLS] embeddings and the output space of the trained linear-probing classifier.

The frozen encoder embeddings form structured regions associated with rhythm labels, qualitatively suggesting that rhythm-relevant information is already present before task-specific adaptation.
After linear probing, several rhythm categories become more clearly separated, while the overall class organization remains qualitatively similar.
This pattern is consistent with the strong linear-probing results in Table~\ref{tab:perform_linear}: the classifier appears to sharpen an already informative representation space, rather than relying entirely on classifier-side separation.

\paragraph{\textbf{Temporal allocation of experts.}}

\begin{figure}[!htbp]
    \vspace{-.2cm}
    \centering
    \includegraphics[width=.99\linewidth]{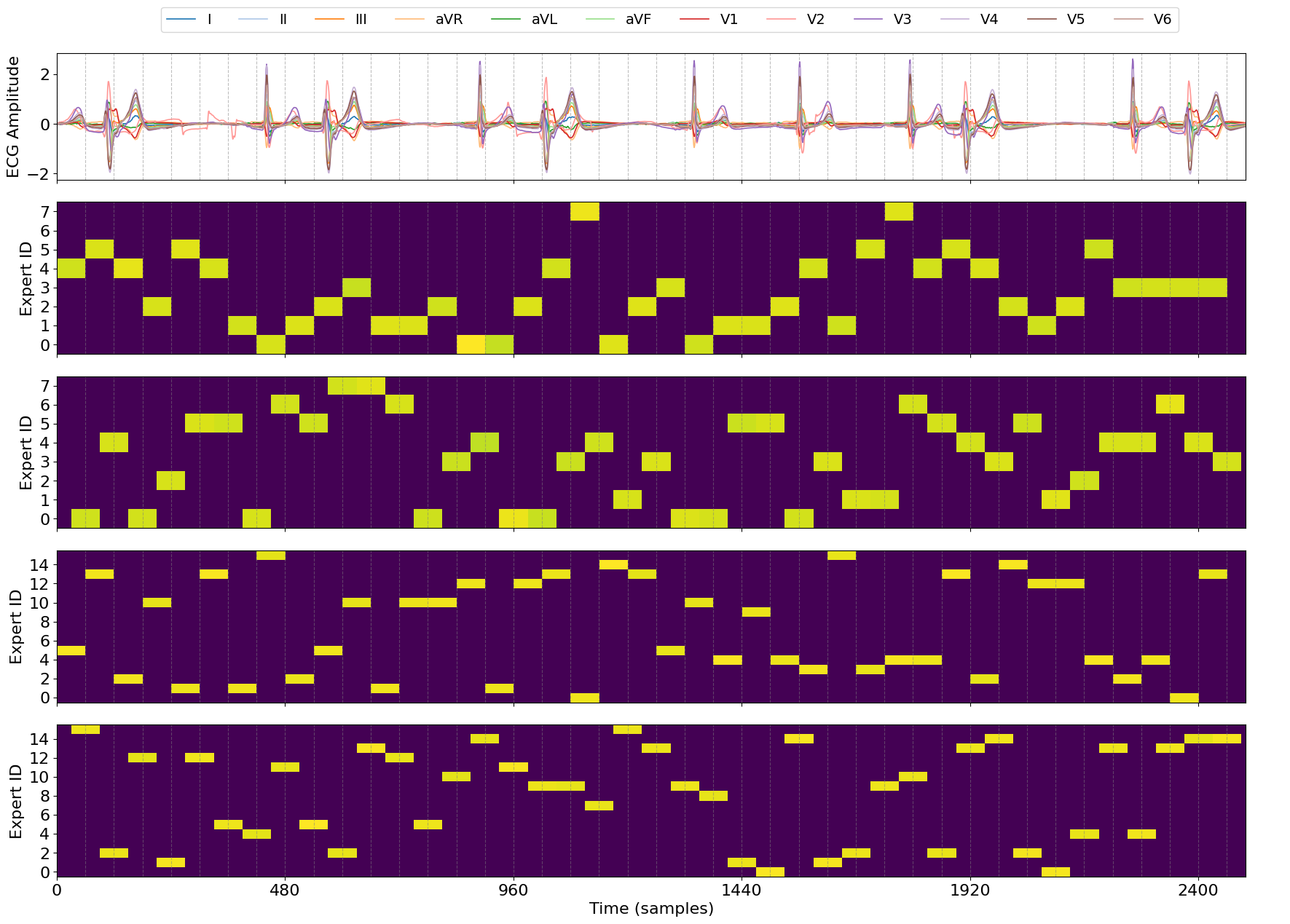}
    \vspace{-.2cm}
    \caption{
    Patch-wise expert allocation for one ECG example.
    The top panel overlays the 12-lead ECG waveform, and the lower panels show the Top-1 expert IDs assigned across overlapping temporal patches for the local- and global-level MoE modules.
    Changes along the temporal axis indicate that different patches or contextual tokens within the same ECG can be routed to different experts.
    }
    \vspace{-.2cm}
    \Description{Example 12-lead ECG waveform with patch-wise expert assignments over time for local and global MoE layers.}
    \label{fig:sample}
\end{figure}

We next visualize how token-level routing operates within an individual ECG recording.
Figure~\ref{fig:sample} shows one 12-lead ECG example together with the Top-1 expert selected for each overlapping temporal patch.
The upper waveform panel provides the temporal ECG context, while the lower allocation panels show how local- and global-level expert IDs change along the same time axis.

The selected experts vary across temporal segments, illustrating that HEART does not assign an entire ECG recording to a single expert.
Instead, different waveform segments and context-integrated tokens can activate different experts within the same sample.
At the local level, routing varies across patch-level waveform segments before temporal contextualization.
At the global level, routing is applied after self-attention, where token representations have already integrated temporal and cross-lead context.
This example illustrates the intended token-level nature of HEART's conditional expert processing.

\paragraph{\textbf{Expert utilization across rhythm labels.}}
\begin{figure}[!htbp]
    \vspace{-.2cm}
    \centering
    \begin{subfigure}[]{\linewidth}
        \centering
        \includegraphics[width=.99\linewidth]{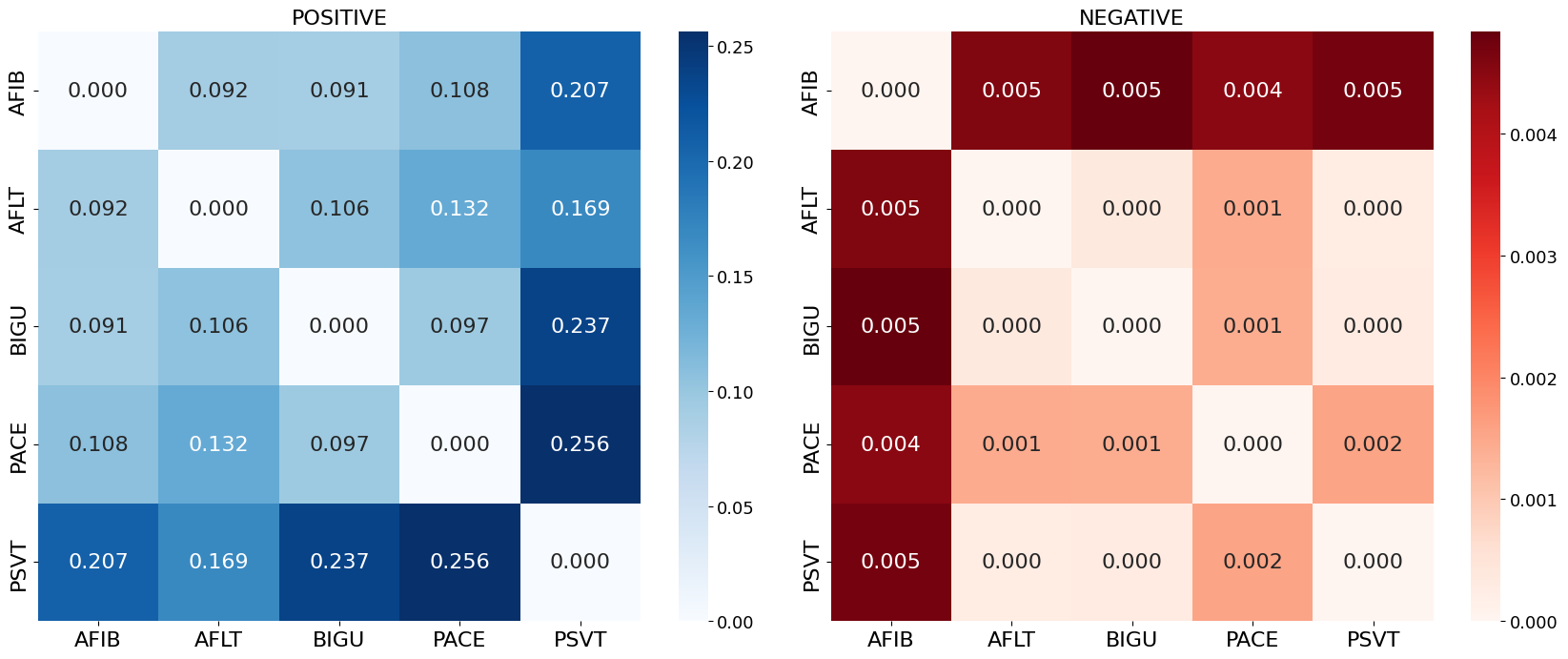}
        \caption{Local-level MoE}
        \label{fig:expert_js_A}
    \end{subfigure}
    \begin{subfigure}[]{\linewidth}
        \centering
        \includegraphics[width=.99\linewidth]{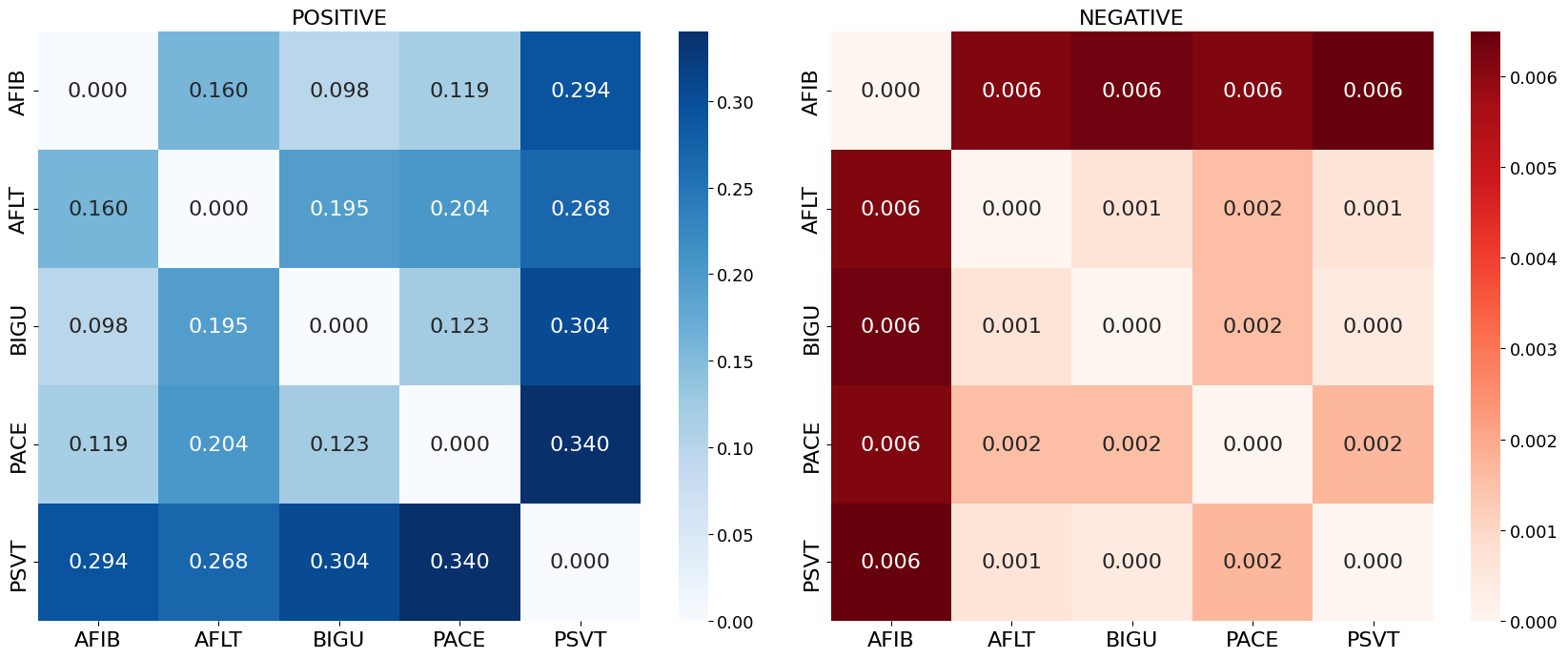}
        \caption{Global-level MoE}
        \label{fig:expert_js_B}
    \end{subfigure}
    \vspace{-.2cm}
    \caption{
    Jensen--Shannon divergence between expert-utilization distributions across PTB-XL Rhythm labels.
    For each target label, positive samples are ECGs annotated with that rhythm label, whereas negative samples are ECGs without that target label.
    Divergence values are computed from patch-aggregated expert-utilization distributions and shown separately for the local- and global-level MoE modules.
    }
    \vspace{-.2cm}
    \Description{Matrices of Jensen-Shannon divergence between expert utilization distributions for rhythm labels, split by local and global MoE levels and by positive and negative samples.}
    \label{fig:expert_js}
\end{figure}

\begin{figure*}[t]
    \centering
    \begin{subfigure}[]{0.99\linewidth}
        \includegraphics[width=\linewidth]{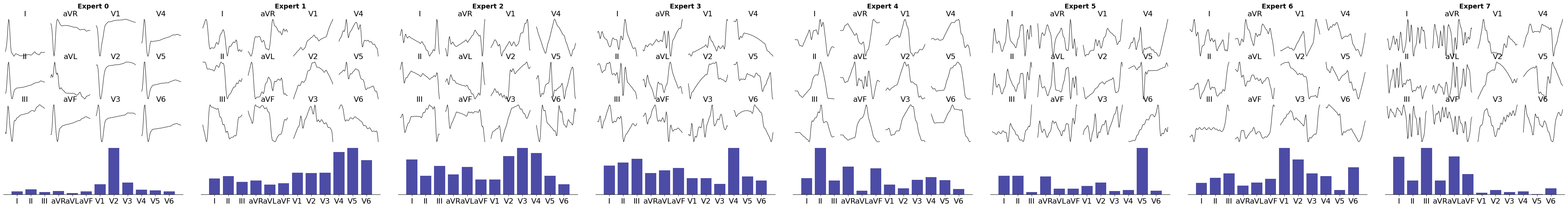}
        \caption{Local-level expert prototypes}
        \label{fig:prototypes1}
    \end{subfigure}
    \begin{subfigure}[]{0.99\linewidth}
        \includegraphics[width=\linewidth]{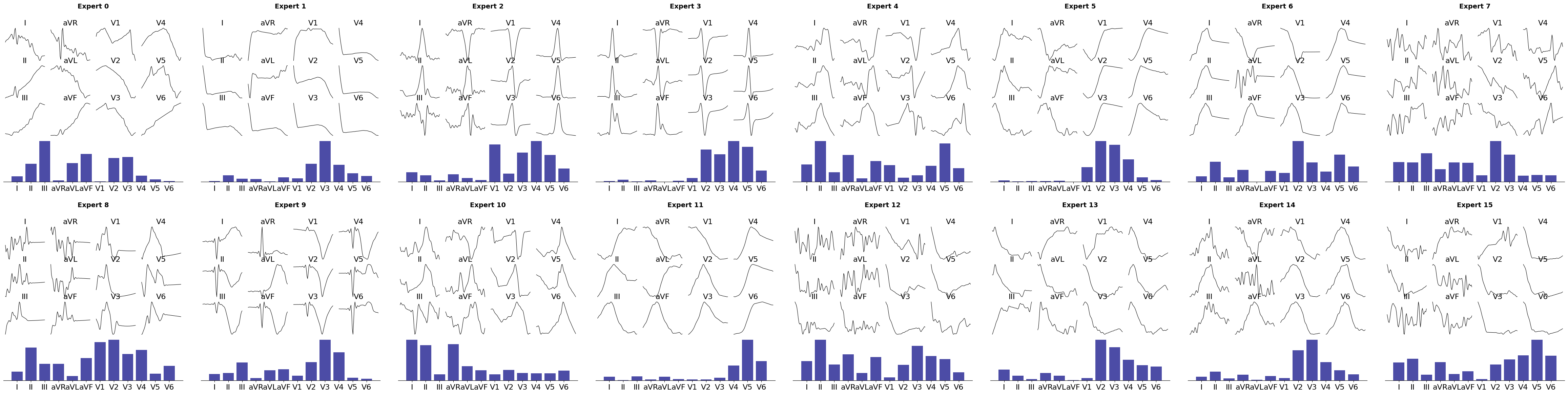}
        \caption{Global-level expert prototypes}
        \label{fig:prototypes2}
    \end{subfigure}
    \vspace{-.3cm}
    \caption{
    Representative ECG patch prototypes for each expert.
    Each panel shows an actual routed ECG patch selected near the expert-specific centroid, together with its normalized per-lead variance profile.
    The prototype is a real ECG segment rather than an averaged waveform, and the bar plot summarizes relative lead-wise variation rather than causal attribution.
    }
    \vspace{-.3cm}
    \Description{Representative ECG patch prototype panels for local and global experts, showing waveform morphology and normalized per-lead variance profiles for each expert.}
    \label{fig:Prototypes_expert}
\end{figure*}

To examine whether expert usage differs across rhythm conditions, we aggregate expert assignments across patches and compare the resulting utilization distributions for each rhythm label.
Figures~\ref{fig:expert_js} and~\ref{fig:hist} (in the Appendix) provide complementary views of the same analysis: the JS divergence heatmaps summarize pairwise differences between label-conditioned expert-utilization distributions, while the histograms show the corresponding expert-wise utilization profiles for each rhythm label.

As shown in Figure~\ref{fig:expert_js}, positive groups exhibit larger JS divergence across rhythm labels than the corresponding negative groups.
This suggests that target-positive rhythm patterns are associated with more label-specific expert-utilization profiles.
The histograms in Figure~\ref{fig:hist} further show how these differences appear at the expert level: local-level utilization is relatively distributed across experts, whereas global-level utilization is more peaked for several rhythm labels.

By contrast, negative groups show near-zero JS divergence across labels, partly because each negative group is defined by the absence of one target rhythm label and can overlap substantially with negative groups for other labels in a multi-label setting.
Thus, the positive--negative contrast is best understood as qualitative support for label-conditioned routing in target-positive samples, without implying that all negative ECGs share a common physiological pattern.

Together, the JS divergence and histogram views are consistent with the intended hierarchy of HEART: local experts tend to exhibit more distributed patch-level usage, while global experts show stronger label-conditioned preferences over context-integrated token representations.

\paragraph{\textbf{Expert-associated morphological prototypes.}}
Finally, we examine expert-associated ECG patterns using representative routed patches.
Each displayed patch contains 60 time points, corresponding to approximately 0.24 seconds of the 250\,Hz ECG signal.
For each expert, we select the routed token closest to the expert-specific centroid in representation space and visualize its corresponding raw ECG patch, so each prototype is a real ECG segment rather than an averaged waveform.
Figure~\ref{fig:Prototypes_expert} shows each selected 12-lead patch together with its normalized per-lead temporal variance profile.
The bar plot summarizes the relative magnitude of temporal variance across leads within the displayed patch.

Local-level prototypes correspond to pre-contextualized patch representations and can therefore be interpreted as short patch-scale waveform patterns associated with each local expert.
Across local experts, the selected prototypes show visible differences in waveform shape, baseline stability, sharp deflections, and noise-related variation, suggesting separation of short-range morphological and signal-quality patterns.
In contrast, global-level prototypes correspond to post-attention tokens; the displayed raw patch is the temporal anchor of a context-integrated representation routed to the expert, rather than the isolated local motif that determined routing.
Across global experts, the anchor patches show distinct waveform motifs and lead-variance profiles, suggesting that context-aware routing is associated with different temporal or cross-lead patterns.
Overall, these observations are consistent with complementary local and global expert behavior, while remaining qualitative rather than causal evidence.

\vspace{-.15cm}
\section{Conclusion}
We presented HEART, a hierarchical ECG foundation model that introduces stage-specific conditional expert refinement into self-supervised ECG representation learning.
By placing MoE modules at both patch-level and context-integrated representation stages, HEART enables token-level specialization within a single ECG recording.
Across PTB-XL and CPSC2018, HEART achieves strong performance under both linear probing and fine-tuning, with particularly strong frozen-encoder results suggesting that downstream-relevant structure is already accessible to a linear classifier.
The comparison with CREMA-Max further suggests that HEART's gains are not explained by dense parameter scaling alone.
Qualitative analyses illustrate patch-wise routing, label-conditioned expert utilization, and expert-associated waveform motifs, providing complementary evidence for local and global expert specialization.
Together, these findings support stage-specific MoE placement as an ECG-motivated design bias for self-supervised representation learning.

\vspace{-.15cm}
\section{Limitations and Future Work}
HEART empirically studies stage-specific MoE refinement within a self-supervised Transformer ECG encoder, but does not theoretically characterize how expert specialization emerges. Some external comparisons rely on reported results due to unavailable implementations or checkpoints and should therefore be interpreted as contextual references. Our qualitative analyses are illustrative rather than causal; future work will extend routing analyses across broader ECG tasks and datasets to better understand expert specialization.


\clearpage
\twocolumn
\bibliographystyle{ACM-Reference-Format}
\bibliography{sample-base}

\clearpage


\appendix

\twocolumn[{
\begin{center}
\captionsetup{type=figure}

\begin{subfigure}{0.95\textwidth}
    \centering
    \includegraphics[width=\linewidth]{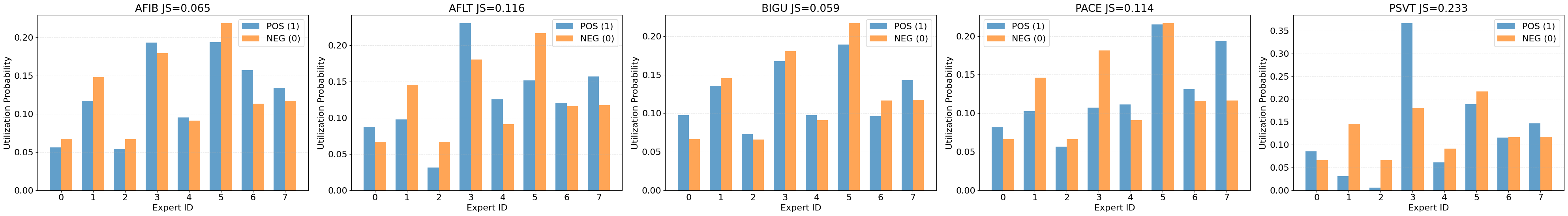}
    \caption{Local-level MoE utilization}
    \label{fig:patch_hist}
\end{subfigure}

\vspace{0.1cm}

\begin{subfigure}{0.95\textwidth}
    \centering
    \includegraphics[width=\linewidth]{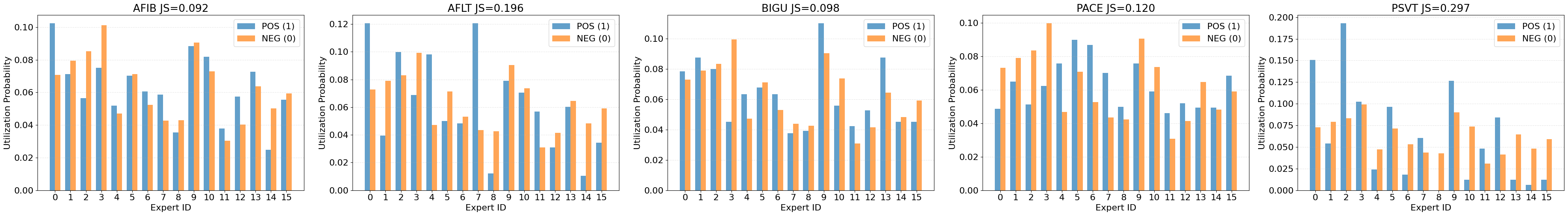}
    \caption{Global-level MoE utilization}
    \label{fig:terminal_hist}
\end{subfigure}

\vspace{-0.1cm}

\captionof{figure}{
Patch-aggregated expert-utilization histograms across PTB-XL Rhythm labels.
Each panel shows how often each expert is selected for positive and negative samples of a target rhythm label, separately for the local- and global-level MoE modules.
}
\Description{Expert utilization histograms for local and global MoE layers across rhythm labels, with separate positive and negative sample distributions.}
\label{fig:hist}

\end{center}
\vspace{0.3cm}
}]

\section{Implementation Details and Hyperparameter Search}
\label{app:implementation}

Table~\ref{tab:hpo_space} summarizes the hyperparameter search space.
The final configuration was selected based on validation performance.

\begin{table}[!htbp]
\centering
\scriptsize
\caption{Hyperparameter search space used for HEART.}
\label{tab:hpo_space}
\begin{tabular}{l l l l}
\toprule
\textbf{Component} & \textbf{Hyperparameter} & \textbf{Distribution} & \textbf{Range / Categories} \\
\midrule
Network & resnet.kernel\_size & choice & $\{3, 7, 15\}$ \\
Network & resnet.dropout & choice & $\{0, 0.1, 0.2\}$ \\
Network & resnet.num\_blocks & choice & $\{2, 4, 8\}$ \\
Network & encoder\_depth & choice & $\{4, 6, 8\}$ \\
Network & encoder\_num\_heads & choice & $\{16, 32\}$ \\
Network & decoder\_depth & choice & $\{4, 6, 8\}$ \\
Network & decoder\_num\_heads & choice & $\{16, 32\}$ \\
\midrule
Local-level MoE & patch\_capacity\_factor & loguniform & $[1.0,\; 2.0]$ \\
Local-level MoE & patch\_router\_z\_loss\_coef & loguniform & $[5\!\times\!10^{-4},\; 5\!\times\!10^{-3}]$ \\
Local-level MoE & patch\_aux\_loss\_coef & loguniform & $[5\!\times\!10^{-3},\; 2\!\times\!10^{-2}]$ \\
Local-level MoE & patch\_jitter\_noise & loguniform & $[10^{-3},\; 10^{-2}]$ \\
\midrule
Global-level MoE & capacity\_factor & loguniform & $[1.0,\; 2.0]$ \\
Global-level MoE & router\_z\_loss\_coef & loguniform & $[5\!\times\!10^{-4},\; 5\!\times\!10^{-3}]$ \\
Global-level MoE & aux\_loss\_coef & loguniform & $[5\!\times\!10^{-3},\; 2\!\times\!10^{-2}]$ \\
Global-level MoE & jitter\_noise & loguniform & $[10^{-3},\; 10^{-2}]$ \\
\midrule
Optimizer (AdamW) & weight\_decay & loguniform & $[10^{-4},\; 10^{-2}]$ \\
Optimizer (AdamW) & lr & loguniform & $[10^{-5},\; 5\!\times\!10^{-4}]$ \\
\bottomrule
\end{tabular}
\end{table}

The selected HEART configuration uses 12-lead ECGs of 10 seconds at 250\,Hz ($T=2500$), divided into patches of size $P=60$ with 50\% overlap, yielding 82 temporal patches.
The patch embedding module consists of a convolutional stem with a kernel size of 15, followed by a lightweight 1D ResNet.
The encoder uses dimension $D=1024$, depth $L=4$, and $H=32$ attention heads.
The local- and global-level MoE modules use 8 and 16 MLP experts, respectively, with GELU activation, RMSNorm, a dropout rate of 0.0, and capacity factors of approximately 1.32 and 1.76.
The contrastive loss weight is set to $\lambda_{\mathrm{cont}}=0.1$; the local auxiliary and router $z$-loss weights are 0.005 and 0.0008, while the corresponding global weights are 0.017 and 0.002.

The raw ECG is processed by the convolutional stem and lightweight 1D ResNet patch embedding module before local-level MoE refinement, while the global-level MoE is applied at the context-integrated encoder stage. Both MoE modules use Top-1 routing with input jitter and noisy gating.

\section{Detailed Routing Objective}
\label{app:moe_routing}

Both local- and global-level MoE modules use the same routing formulation.
For token representation $\mathbf{h}_n\in\mathbb{R}^{D}$, the router computes base logits with optional input jitter:
{\small
\begin{align*}
    \mathbf{u}_n 
    &= W_g(\mathbf{h}_n+\boldsymbol{\epsilon}_n),
    &
    \boldsymbol{\epsilon}_n 
    &\sim \mathcal{U}(-\sigma,\sigma).
\end{align*}
}

With noisy gating, the routing logits, probabilities, and Top-1 assignment are defined as:
{\small
\begin{align*}
    \mathbf{r}_n 
    &= \mathbf{u}_n + \boldsymbol{\eta}_n,
    &
    \boldsymbol{\eta}_n 
    &\sim \mathcal{N}(0,\Sigma(\mathbf{h}_n)), \\
    \mathbf{p}_n 
    &= \mathrm{Softmax}(\mathbf{r}_n/\tau),
    &
    \hat{a}_n 
    &= \arg\max_i p_{n,i}.
\end{align*}
}

Each expert processes at most
{\small
\begin{align*}
    C_{\mathrm{cap}}
    =
    \left\lceil
    \frac{N}{E}\gamma
    \right\rceil
\end{align*}
}
tokens, where $N$ is the number of routed tokens, $E$ is the number of experts, and $\gamma$ is the capacity factor.
Let $m_{n,i}\in\{0,1\}$ denote the final dispatch mask after capacity filtering.
For expert $i$, the dispatched token fraction, mean routing probability, and load-balancing loss are:
{\small
\begin{align*}
    f_i
    &= \frac{1}{N}\sum_{n=1}^{N}m_{n,i},
    &
    P_i
    &= \frac{1}{N}\sum_{n=1}^{N}p_{n,i},
    &
    \mathcal{L}_{\mathrm{bal}}
    &=
    E\sum_{i=1}^{E}f_iP_i .
\end{align*}
}

The auxiliary routing objective further includes routing entropy and router logit regularization:
{\small
\begin{align*}
    \mathcal{L}_{\mathrm{ent}}
    &=
    -
    \frac{1}{N}
    \sum_{n=1}^{N}
    \sum_{i=1}^{E}
    p_{n,i}\log(p_{n,i}+\epsilon),
    &
    \mathcal{L}_{z}
    &=
    \frac{1}{N}
    \sum_{n=1}^{N}
    \|\mathbf{u}_n\|_2^2 .
\end{align*}
}

Here, $\mathcal{L}_{z}$ is computed on the base router logits before Gaussian noise and temperature scaling.
For one MoE module, the auxiliary loss is:
{\small
\begin{align*}
    \mathcal{L}_{\mathrm{aux}}
    &=
    \mathcal{L}_{\mathrm{bal}}
    -
    \alpha\mathcal{L}_{\mathrm{ent}}
    +
    \lambda_z\mathcal{L}_{z},
    &
    \mathcal{L}_{\mathrm{aux}}^{\mathrm{HEART}}
    &=
    \mathcal{L}_{\mathrm{aux}}^{\mathrm{local}}
    +
    \mathcal{L}_{\mathrm{aux}}^{\mathrm{global}} .
\end{align*}
}

The auxiliary routing losses are used only during self-supervised pre-training and are inserted into the total objective in Eq.~\ref{eq:total_loss}.

\end{document}